%% file: ls.tex
\documentclass{article}
\usepackage{times}
\usepackage{amssymb}
\usepackage{soul}
\usepackage{url}
\usepackage{graphicx}
\usepackage{multicol}
\usepackage{amsmath} 

\usepackage{amsfonts}
\usepackage{subcaption}
\usepackage{comment} 
\usepackage{lipsum} 
\usepackage{fullpage} 
\usepackage{utopia} 
\usepackage{etoolbox}
\usepackage{array}
\usepackage{booktabs}
\usepackage{algorithm}
\usepackage{algpseudocode}
\usepackage{tabu}
\usepackage{thmtools,thm-restate}

\algblock{Input}{EndInput}
\algnotext{EndInput}
\algblock{Output}{EndOutput}
\algnotext{EndOutput}

\newtheorem{theorem}{Theorem}

\newtheorem{corollary}[theorem]{Corollary}
\newtheorem{lemma}[theorem]{Lemma}
\newtheorem{definition}{Definition}

\newcommand{\algmargin}{\the\ALG@thistlm}

\input{lib/symbol}
\input{lib/consistency}
\newlength{\whilewidth}
\algdef{SE}[parWHILE]{parWhile}{EndparWhile}[1]
{\parbox[t]{\dimexpr\linewidth-\algmargin}{%
		\hangindent\whilewidth\strut\algorithmicwhile\ #1\ \algorithmicdo\strut}}{\algorithmicend\ \algorithmicwhile}%
\algnewcommand{\parState}[1]{\State%
	\parbox[t]{\dimexpr\linewidth-\algmargin}{\strut #1\strut}}
\usepackage{multicol}
\usepackage{MnSymbol,wasysym}
\usepackage{graphicx}
\usepackage{color}
\usepackage{subcaption}
\usepackage{verbatim}
\usepackage{bm}
\usepackage[colorlinks,linkcolor=blue,anchorcolor=blue,citecolor=blue,]{hyperref}
\usepackage{tgpagella}
\usepackage{subfiles}
\usepackage{subcaption}
\usepackage{mwe}
\usepackage{multirow}
\usepackage{tabu}
\usepackage{makecell}
\usepackage{epsfig}
\usepackage{advdate} 
\usepackage{calc}
\usepackage{yfonts}

\usepackage[square]{natbib}
\usepackage{tikz}

\newcommand{\OT}{\operatorname{OT}}
\newcommand{\LOT}{\operatorname{LOT}}
\newcommand{\LSOT}{\operatorname{LSOT}}
\newcommand{\rankp}{\operatorname{rank}_+}
\newcommand{\yp}{{\mathbf{y}^{\mathbf{p}}}}
\newcommand{\yq}{{\mathbf{y}^{\mathbf{q}}}}

\newcommand{\ia}{I^{\mathbf{A}}}
\newcommand{\ib}{I^{\mathbf{B}}}
\newcommand{\is}{I^{\mathbf{S}}}

\newcommand{\vecto}{\operatorname{vec}}
\newcommand{\quo}{//}
\newcommand{\rem}{\%}
\newcommand{\dist}{\operatorname{dist}}
\newcommand{\bcd}{\operatorname{bcd}}
\newcommand{\lag}{\mathfrak{L}}
\newcommand{\con}{\alpha}
\newcommand{\conB}{\alphaB}
\newcommand{\clamp}{\operatorname{clamp}}
\newcommand{\shrin}{\operatorname{shrin}}
\newcommand{\onem}{{\oneB^m}}
\newcommand{\onen}{{\oneB^n}}

\title{Approximating Optimal Transport via Low-rank and Sparse Factorization}
\author {

        Weijie Liu,\textsuperscript{\rm 1}
        Chao Zhang, \textsuperscript{\rm 1}
        Nenggan Zheng, \textsuperscript{\rm 1}
        Hui Qian \textsuperscript{\rm 1}\\
}
\date{%
	\textsuperscript{\rm 1} Zhejiang University \\%
	\{westonhunter, zczju, zng, qianhui\}@zju.edu.cn\\[2ex]%
	\today
}

\begin{document}

\maketitle

\begin{abstract}
	Optimal transport (OT) naturally arises in a wide range of machine learning applications but may often become the computational bottleneck.
	Recently, one line of works propose to solve OT approximately by searching the \emph{transport plan} in a low-rank \subspace.
	However, the optimal transport plan is often not low-rank, which tend to yield large approximation errors.
	For example, when Monge's \emph{transport map} exists, the induced transport plan is full rank.
	This paper concerns the computation of the OT distance with adequate accuracy and efficiency.
	A novel approximation for OT is proposed, in which the transport plan can be decomposed into the sum of a low-rank matrix and a sparse one.
	We theoretically analyze the approximation error.
	An augmented Lagrangian method is then designed to efficiently calculate the transport plan.
\end{abstract}


\section{Introduction}
\subfile{sections/introduction.tex}

\section{Preliminaries}
\subfile{sections/preliminaries.tex}

\section{Methodology}
\subfile{sections/methodology.tex}

\section*{Conclusion}

In this paper, we propose a novel approximation of the OT distance.
The optimal transport plan is approximated by the sum of a low-rank matrix and a sparse one.
An augmented Lagrangian method is designed to efficiently calculate the transport plan.

\bibliographystyle{plainnat}
\bibliography{ls}

\clearpage
\appendix
\section{Omitted Proofs}
\subfile{sections/proof.tex}

\end{document}

%% file: lib/symbol.tex
\newcommand{\AB}{\mathbf{A}}
\newcommand{\BB}{\mathbf{B}}
\newcommand{\CB}{\mathbf{C}}
\newcommand{\DB}{\mathbf{D}}

\newcommand{\HB}{\mathbf{H}}

\newcommand{\KB}{\mathbf{K}}
\newcommand{\LB}{\mathbf{L}}
\newcommand{\MB}{\mathbf{M}}

\newcommand{\RB}{\mathbf{R}}
\newcommand{\SB}{\mathbf{S}}
\newcommand{\TB}{\mathbf{T}}

\newcommand{\XB}{\mathbf{X}}

\newcommand{\ZB}{\mathbf{Z}}

\newcommand{\zeroB}{\mathbf{0}}
\newcommand{\oneB}{\mathbf{1}}
\newcommand{\aB}{\mathbf{a}}
\newcommand{\bB}{\mathbf{b}}

\newcommand{\fB}{\mathbf{f}}
\newcommand{\gB}{\mathbf{g}}

\newcommand{\pB}{\mathbf{p}}
\newcommand{\qB}{\mathbf{q}}

\newcommand{\uB}{\mathbf{u}}
\newcommand{\vB}{\mathbf{v}}

\newcommand{\xB}{\mathbf{x}}
\newcommand{\yB}{\mathbf{y}}

\newcommand{\RBB}{\mathbb{R}}

\newcommand{\EBB}{\mathbb{E}}

\newcommand{\BM}{\mathcal{B}}

\newcommand{\IM}{\mathcal{I}}

\newcommand{\OM}{\mathcal{O}}

\newcommand{\XM}{\mathcal{X}}

\newcommand{\alphaB}{\mbox{\boldmath$\alpha$\unboldmath}}

\newcommand{\argmin}{\mathop{\rm argmin}}
\newcommand{\rank}{\mathop{\rm rank}}

\newcommand{\diag}{\mathsf{diag}}

\newcommand{\trace}{\operatorname{trace}}
\newcommand{\KL}{\operatorname{KL}}
\newcommand{\proj}{\operatorname{Proj}^{\KL}_{\Pi(\pB,\qB)}}

\newcommand{\op}{\operatorname{op}}

%% file: lib/consistency.tex
\newcommand{\bilinear}{bi-linear}

\newcommand{\elementwise}{elementwise}

\newcommand{\largescale}{large-scale}

\newcommand{\nonconvex}{non-convex}
\newcommand{\nonnegative}{non-negative}
\newcommand{\nonnegativity}{non-negativity}
\newcommand{\nonzero}{nonzero}
\newcommand{\realworld}{real-world}

\newcommand{\subproblem}{sub-problem}
\newcommand{\subproblems}{sub-problems}

\newcommand{\subroutine}{sub-routine}
\newcommand{\subspace}{sub-space}

\newcommand{\Nonnegative}{Non-negative}

%% file: sections/introduction.tex
Optimal transport (OT) defines the distance between two probability measures \citep{villani2009optimal}.
It has a wide range of machine learning applications, including generative modeling \citep{arjovsky2017wasserstein}, domain adaptation \citep{courty2016optimal}, and data mining \citep{xu2019gromov}, to name but a few.


Despite the broad applications, calculating the OT distance easily becomes the computational bottleneck in \realworld{} problems.
Originally, the OT problem was solved via linear programming, which involves the formidable computational complexity $\OM(n^3\log n)$ where $n$ is the size of discrete measures \citep{tarjan1997dynamic}.
A popular method, known as Sinkhorn's method, regularizes the \emph{transport plan} with its entropy and accelerates the optimization of the transport plan \citep{cuturi2013sinkhorn}.
Given a data-dependent cost matrix $\CB$, each iteration of Sinkhorn's method takes the form of matrix-vector products $\exp(\frac{-\CB}{\eta})\uB$ or $\exp(\frac{-\CB}{\eta})^\top\vB$ where $\eta$ is the weight of the entropy regularizer and $\uB$, $\vB$ are $n$-dimensional vectors.
Such an approach can obtain an $\epsilon$-approximation of the OT distance with complexity $\OM(\frac{n^2\log n}{\epsilon^3})$, which is still computationally expensive when a highly accurate solution is required.

{\bf Low-rank kernel factorization.}
Recently, one line of works speed up the Sinkhorn's method  by using a rank-$r$ approximation of $\exp(\frac{-\CB}{\eta})$ \citep{altschuler2019massively,altschuler2020polynomial,scetbon2020linear}.
Adopting this approximation, the cost for each iteration of the Sinkhorn's method can be reduced to $\OM(nr^2)$.
However, to approximate $\exp(\frac{-\CB}{\eta})$ with sufficient accuracy, $r$ can be large \citep{altschuler2019massively}.
When $r^2$ is close to or ever larger than $n$, these methods can hardly yield improved performance over Sinkhorn's method.

{\bf Low-rank transport plan.}
Another family of works accelerate the calculation by searching the transport plan in a low-rank \subspace{} \citep{forrow2019statistical,lin2021making,scetbon2021low}.
However, the optimal transport plan may not be low-rank, which leads to poor approximation of these methods.
To illustrate this, consider a setting where the target probability measure is obtained by permuting the supports of the source measure.
In such a case, the optimal transport plan is a permutation matrix which is full-rank.

In this paper, we propose a novel approximation for the OT distance in which the transport plan is in a new \subspace{}.
The transport plan can be decomposed into the sum of a low-rank matrix and a sparse one.
An inexact augmented Lagrangian method is designed to efficiently resolve the resulted optimization problem by solving a series of \subproblems{}.
We handle each \subproblem{} via a block coordinate descent \subroutine{}.
Our contributions are summarized as follows.
\begin{enumerate}
	\item We theoretically analyze the error of the proposed approximation.
	\item We propose an inexact augmented Lagrangian method to calculate the transport plan.
\end{enumerate}


\paragraph{Notation.}
We use bold lowercase symbols (e.g. $\xB$), bold uppercase letters (e.g. $\XB$), uppercase calligraphic fonts (e.g. $\XM$), and Greek letters (e.g. $\alpha$), to denote vectors, matrices, spaces (sets), and measures, respectively.
$\oneB^d\in\RBB^d$ and $\zeroB^d\in\RBB^d$ are the all-ones vector and the all-zeros vector respectively, where $\RBB^d$ is the $d$-dimensional Euclidean space.
$\RBB_+^d$ is the subspace of $\RBB^d$ and contains \nonnegative{} entries.
$\xB\ge c$ (resp. $\XB\ge c$) means each element of vector $\xB$ (resp. matrix $\XB$) is greater than or equal to scalar $c$.
Given a matrix $\XB$, we denote by $\|\XB\|_F$ its Frobenius norm, by $\|\XB\|_0$ its number of \nonzero{} entries, by $\|\XB\|_1$ its \elementwise{} $\ell_1$ norm (i.e., $\|\XB\|_1=\sum_{ij}|X_{ij}|$), and by $\|\XB\|_\infty$ its \elementwise{} $\ell_\infty$ norm (i.e., $\|\XB\|_\infty=\max_{ij}|X_{ij}|$).
For two matrices $\AB$ and $\BB$ that are of the same size, $\langle \AB,\BB\rangle=\trace(\AB^\top\BB)$ is the Frobenius dot-product.
$(\aB;\bB)$ denotes the concatenation of vectors $\aB$ and $\bB$.
The vectorization of matrix $\XB$ (in the row order) is denoted by $\vecto(\XB)$.
A discrete measure $\alpha$ can be denoted by $\alpha=\sum_{i=0}^{m-1}p_i\delta_{\xB_i}$ where $\delta_{\xB}$ is the Dirac at position $\xB$, i.e., a unit of mass infinitely concentrated at $\xB$.
With slight abuse of notation, we also use $\pB=[p_i]$ to refer to $\alpha$.




%% file: sections/preliminaries.tex
%


\subsection{Optimal Transport and Sinkhorn Method}
The OT distance \citep{villani2009optimal,cuturi2013sinkhorn} between discrete measures $\pB$ and $\qB$ is defined as
\begin{equation}\label{eq:unregularized_OT}
	\OT(\pB,\qB)=\min_{\TB\in\Pi(\pB,\qB)}\langle\CB,\TB\rangle,
\end{equation}
where the $(i,j)$\textsuperscript{th} entry of $\CB$ is the distance between the $i$\textsuperscript{th} support of $\pB$ and the $j$\textsuperscript{th} support of $\qB$, and the feasible domain of transport plan $\TB=[T_{ij}]$ is given by the set $\Pi(\pB,\qB)=\{\TB\in\RBB_+^{m\times n}|\TB\oneB^n=\pB,\TB^\top\oneB^m=\qB\}$.

\citeauthor{cuturi2013sinkhorn}~\citeyear{cuturi2013sinkhorn} proposes to solve the following the entropy regularized OT problem,
\begin{equation*}
	\min_{\TB\in\Pi(\pB,\qB)}\langle\CB,\TB\rangle-\eta H(\TB),
\end{equation*}
where $H(\TB)$ is the entropy of the transport plan, i.e., $H(\TB)=-\sum_{i=0}^{m-1}\sum_{j=0}^{n-1}T_{ij}(\log T_{ij}-1)$.
By setting $\eta=\OM(\frac{\epsilon}{\log mn})$, the Sinkhorn's method computes an $\epsilon$-approximate solution of the problem (\ref{eq:unregularized_OT}) in $\OM(\frac{n^2\log mn}{\epsilon^3})$ operations \citep{altschuler2017near}, which may still be too expensive for \largescale{} problems especially when a highly accurate solution of (\ref{eq:unregularized_OT}) is required.

\subsection{\Nonnegative{} factorization of the Transport Plan}
\citeauthor{scetbon2021low}~\citeyear{scetbon2021low} force the transport plan to be low-rank by using the notion of the \nonnegative{} rank which is formally defined as follows.
\begin{definition}\label{def:nonnegative}
	The \nonnegative{} rank of matrix $\MB$ is the smallest number of \nonnegative{} rank-one matrices into which $\MB$ can be decomposed additively, i.e.,
	\begin{equation*}
		\rankp(\MB)=\min\{q|\MB=\sum_{i=0}^{q-1}\RB_i,\forall i, \rank(\RB_i)=1,\RB_i\ge0\}.
	\end{equation*}
\end{definition}
Specifically, they consider problem
\begin{equation}
	\LOT_{r}(\pB,\qB)=\min_{\TB\in\Pi_r(\pB,\qB)}\langle\CB,\TB\rangle,
\end{equation}
where $\Pi_r(\pB,\qB)=\{\TB\in\Pi(\pB,\qB)|\rankp(\TB)\le r\}$.

From Definition \ref{def:nonnegative}, one has $\rank(\TB)\le\rankp(\TB)\le r$ for all $\TB\in\Pi_r(\pB,\qB)$.
However, the optimal transport plan is often not low-rank, as we have explained in Introduction.

\subsection{Low-rank and Sparse Decomposition}
Approximating a matrix by the sum of a low-rank matrix and a sparse matrix has a long history \citep{candes2011robust}.
Mathematically, such decomposition can be formulated as the following optimization problem
\begin{equation*}
	\min_{\LB,\SB}\rank(\LB)+\lambda\|\SB\|_0, \text{ s.t. } \LB+\SB=\MB.
\end{equation*}
Such \nonconvex{} problem is computationally intractable.
One common approach is to solve the following surrogate problem which has a convex objective (see, e.g. \citet{wright2009robust,lin2011linearized}),
\begin{equation*}
	\min_{\LB,\SB}\|\LB\|_*+\lambda\|\SB\|_1, \text{ s.t. } \LB+\SB=\MB,
\end{equation*}
where the nuclear norm and the $\ell_1$ norm induce the sparsity of the singular values of $\LB$ and the entries of $\SB$ respectively.
As we shall see shortly, the low-rank and sparse decomposition yields better approximation for the OT distance (\ref{eq:unregularized_OT}) than the sole low-rank component.

%% file: sections/methodology.tex
We first define the approximated distance which decomposes the transport plan into a low-rank matrix and a sparse one.
Next, we derive an augmented Lagrangian method for calculating the proposed distance.
Finally, we analyze the overall complexity of the proposed method.

\subsection{Approximation of OT Distance}\label{sec:approximated_distance}
Given \nonnegative{} integers $r\ll\min\{m,n\}$ and $\rho\ll mn$ that control the rank and the sparsity respectively, the proposed approximation of OT distance is
\begin{equation}\label{eq:formuation}
	\LSOT_{r,\rho}(\pB,\qB)=\min_{\TB\in\tilde\Omega_{r,\rho}(\pB,\qB)}\langle \CB,\TB\rangle,
\end{equation}
where the feasible domain is defined as
\begin{equation*}
	\begin{aligned}
		\tilde\Omega_{r,\rho}(\pB,\qB)=\Big\{\TB\in\Pi(\pB,\qB)&\Big|\TB=\LB+\SB,\LB\ge 0,\SB \ge 0, \rankp(\LB) \le r, \|\SB\|_0 \le \rho\Big\}.
	\end{aligned}
\end{equation*}
We theoretically analyze the approximation error of $\LSOT_{r,\rho}(\pB,\qB)$ in the theorem below.

\begin{restatable}{theorem}{primetheorem}\label{thm:ls}
	Denote $\TB^*\in\argmin_{\TB\in\Pi(\pB,\qB)}\langle \CB,\TB\rangle$.
	Assume there exist $\LB^*\in\RBB_+^{m\times n}$ and $\SB^*\in\RBB^{m\times n}$ such that $\TB^*=\LB^*+\SB^*$, $\rankp(\LB^*)=r^*$, and $\|\SB^*\|_0=\rho^*$.
	Let $\tilde{\LB^*}$ and $\tilde{\SB^*}$ be the best approximations of $\LB^*$ and $\SB^*$ respectively (in terms of the Frobenius norm) satisfying $\rankp(\tilde{\LB^*}) \le r$ and $\|\tilde{\SB^*}\|_0\le \rho$.
	Then,
	\begin{equation}\label{eq:ls}
		\LSOT_{r,\rho}(\pB,\qB)-\OT(\pB,\qB) \le \frac{U\|\CB\|_\infty }{\delta}\big(\sqrt{mn}(r^*-r)_++(\rho^*-\rho)_+\big),
	\end{equation}
	where $U=\max\{\|\LB^*\|_\infty,\|\SB^*\|_\infty\}$ and
	\begin{equation}\label{eq:delta}
		\delta=\frac{1}{e^2}\min\big\{\min_{i,j}\{L^*_{ij}+S^*_{ij}|L^*_{ij}+S^*_{ij}>0\},\min_{i,j}\{\tilde{L^*_{ij}}+\tilde{S^*_{ij}}|\tilde{L^*_{ij}}+\tilde{S^*_{ij}}>0\}\big\}.
	\end{equation}
\end{restatable}
The proof is deferred to the appendix.
We further have the two following corollaries.

\begin{corollary}\label{cor:sparsity}
	$\LSOT_{r,\rho}(\pB,\qB)$ recovers $\OT(\pB,\qB)$ with $\rho\ge m+n-1$, i.e.,
	\begin{equation*}
		\LSOT_{r,\rho}(\pB,\qB)=\OT(\pB,\qB),\forall r\ge 0, \text{ and }\rho\ge m+n-1.
	\end{equation*}
\end{corollary}
This is the direct result of Theorem \ref{thm:ls} and the fact that OT distance can be achieved with the transport plan containing up to $m+n-1$ \nonzero{} entries \citep{brualdi2006combinatorial}.
It is generally difficult to determine the \nonnegative{} rank of an optimal transport plan.
However, $\OT(\pB,\qB)$ can be recovered with $\rho\ge m+n-1$.

\begin{corollary}\label{cor:lowrank_OT}
	Denote $\TB^*\in\argmin_{\TB\in\Pi(\pB,\qB)}\langle \CB,\TB\rangle$.
	Assume there exist $\LB^*\in\RBB_+^{m\times n}$ and $\SB^*\in\RBB^{m\times n}$ such that $\TB^*=\LB^*+\SB^*$, $\rankp(\LB^*)=r^*$, and $\|\SB^*\|_0=\rho^*$.
	Let $\tilde{\LB^*}$ be the best approximations of $\LB^*$  (in terms of the Frobenius norm) satisfying $\rankp(\tilde{\LB^*}) \le r$.
	Then,
	\begin{equation}\label{eq:lowrank_OT}
		\LOT_{r}(\pB,\qB)-\OT(\pB,\qB) \le \frac{U\|\CB\|_\infty }{\delta_1}\big(\sqrt{mn}(r^*-r)_++\rho^*\big),
	\end{equation}
	where $U=\max\{\|\LB^*\|_\infty,\|\SB^*\|_\infty\}$ and
	\begin{equation}\label{eq:delta1}
		\delta_1=\frac{1}{e^2}\min\big\{\min_{i,j}\{L^*_{ij}+S^*_{ij}|L^*_{ij}+S^*_{ij}>0\},\min_{i,j}\{\tilde{L^*_{ij}}|\tilde{L^*_{ij}}>0\}\big\}.
	\end{equation}
\end{corollary}
Setting $\rho=0$, $\LOT_{r}(\pB,\qB)$ can obviously be recovered by $\LSOT_{r,\rho}(\pB,\qB)$.
As is stated in Theorem \ref{thm:ls} and Corollary \ref{cor:lowrank_OT}, $\LSOT_{r,\rho}(\pB,\qB)$ approximates $\OT(\pB,\qB)$ better than $\LOT_{r}(\pB,\qB)$, if $\rho^*>0$ for all $\TB^*$.
%

\subsection{Optimization}\label{sec:optimization}

\paragraph{Surrogate problem.}
The $\|\SB\|_0\le \rho$ constraint in $\tilde\Omega_{r,\rho}(\pB,\qB)$ makes problem (\ref{eq:formuation}) intractable.
We thus consider the following surrogate problem
\begin{equation}\label{eq:surrogate}
	\min_{\AB,\BB,\SB\in\Omega_{r,\rho}(\pB,\qB)}\langle \CB,\AB\BB^\top+\SB\rangle+\lambda\|\SB\|_1,
\end{equation}
where the feasible domain is given by
\begin{equation}\label{eq:each_constraint}
	\begin{aligned}
		\Omega_{r,\rho}(\pB,\qB)=\Big\{\AB\in\RBB^{m\times r},\BB\in\RBB^{n\times r},\SB\in\RBB^{m\times n}&\Big|0\le\AB\le 1,0\le\BB\le 1,0\le\SB\le 1,\\
		&(\AB\BB^\top+\SB)\oneB=\pB,(\AB\BB^\top+\SB)^\top\oneB=\qB\Big\}.
	\end{aligned}
\end{equation}
Because of the \bilinear{} terms, (\ref{eq:surrogate}) has a \nonconvex{} objective function and \nonconvex{} constraints.
Stationary points can be found effectively by an inexact augmented Lagrangian method (ALM).

\paragraph{Inexact ALM.}
ALM is a classical algorithm for constrained optimization \citep{hestenes1969multiplier,powell1969method}.
For solving (\ref{eq:surrogate}), ALM suggests solving
\begin{equation}\label{eq:problem_alm}
	\min_{\AB,\BB,\SB}\max_{\yp,\yq}\lag(\AB,\BB,\SB,\yp,\yq,\beta)+\lambda\|\SB\|_1+\ia(\AB)+\ib(\BB)+\is(\SB),
\end{equation}
where $\ia(\cdot)$, $\ib(\cdot)$ and $\is(\cdot)$ are indicator functions corresponding to $0\le\AB\le 1$, $0\le\BB\le 1$, and $0\le\SB\le 1$ respectively.
The augmented Lagrangian function is defined as
\begin{equation}
	\begin{aligned}
		\lag(\AB,\BB,\SB,\yp,\yq,\beta)&=\langle\CB,\AB\BB^\top+\SB\rangle+\frac{\beta}{2}\Big(\|(\AB\BB^\top+\SB)\oneB-\pB\|^2+\|(\AB\BB^\top+\SB)^\top\oneB-\qB\|^2\Big)\\
		&+\langle \yp,(\AB\BB^\top+\SB)\oneB-\pB\rangle+\langle \yq,(\AB\BB^\top+\SB)^\top\oneB-\qB\rangle,
	\end{aligned}
\end{equation}
where $\yp\in\RBB^m$ and $\yq\in\RBB^n$ are multiplier variables, and $\beta>0$ is the penalty parameter.
In the algorithm and the later analysis, we use some additional notation.
For ease of notation, variables $\AB$, $\BB$, and $\SB$ are sometimes referred to as $\xB$, where $\xB=\big(\vecto(\AB);\vecto(\BB);\vecto(\SB)\big)$.
Similarly, $\yB=\big(\yp;\yq\big)$.
With slight abuse of notation, we use $\lag(\AB,\BB,\SB,\yp,\yq,\beta)$ and $\lag(\xB,\yB,\beta)$ interchangeably.
Based on $\xB$, the equality constraints in $\Omega_{r,\rho}(\pB,\qB)$ can be rewritten as $\conB(\xB)=\zeroB$, where $\conB:\XM\to\RBB^{m+n}$ is a vector function with the $u$\textsuperscript{th} entry given by
\begin{equation*}
	\con_u(\xB)=\begin{cases}
		&\sum_{j=0}^{n-1}(\sum_{k=0}^{r-1}A_{uk}B_{jk}+S_{uj})-p_u,\text{ if }0\le u<m,\\
		&\sum_{i=0}^{m-1}(\sum_{k=0}^{r-1}A_{ik}B_{u-m,k}+S_{i,u-m})-q_{u-m}, \text{ otherwise.}
	\end{cases}
\end{equation*}
We further denote
\begin{equation*}
	h(\xB)=\lambda\|\SB\|_1+\ia(\AB)+\ib(\BB)+\is(\SB).
\end{equation*}
The pseudocode of the proposed inexact ALM is demonstrated in Algorithm \ref{alg:ialm}.
The high-level intuition is that we construct a series of strongly-convex and smooth \subproblems{}, each of which is solved inexactly by the block coordinate descent (BCD) in Algorithm \ref{alg:bcd}.
Such an approach is guaranteed by the following proposition.
The proof is provided in the appendix.
\begin{restatable}{proposition}{propsmoothsc}\label{prop:smooth_sc}
	Given $\yB$ and $\beta$, let $G(\xB)=\lag(\xB,\yB,\beta)+L(\yB,\beta)\|\xB-\bar{\xB}\|^2$ where
	\begin{equation}\label{eq:smoothness}
		L(\yB,\beta)=\sqrt{2r}\|\CB\|_F+\sqrt{2r}(m\sqrt{n}+n\sqrt{m})\|\yB\|+\beta L_c,
	\end{equation}
	with
	\begin{equation}
		\begin{aligned}
			B_u&=\max_{\xB\in\XM}\max\{|\con_u(\xB)|,\|\nabla \con_u(\xB)\|\},\\
			L_c&=\sum_{u=0}^{m-1}\sqrt{2nr}B_u+\sum_{u=m}^{m+n-1}\sqrt{2mr}B_u+\sum_{u=0}^{m+n-1}B_u^2.
		\end{aligned}
	\end{equation}
	Then, $G(\xB)$ is $3L(\yB,\beta)$-smooth and $L(\yB,\beta)$-strongly convex.
\end{restatable}

\begin{algorithm}[ht]
	\caption{Inexact augmented Lagrangian method\label{alg:ialm}}
	\begin{algorithmic}[1]
		\State {\textbf{Input:}} $\epsilon$, $\beta_0>0$, $\sigma>1$, $w_0$, and $T$.
		\State {\textbf{Output:}} $\xB_T$.
		\State {\textbf{Initialization:}} $\xB_0\in\XM$, $\yB_0=\zeroB$.
		\For{$t=0$, $1$, $\dots$, $T-1$}
		\State Calculate $\beta_t=\beta_0\sigma^t$ and $L_t=L(\yB_t,\beta_t)$ as (\ref{eq:smoothness}).
		\State $\xB_{0,t}=\xB_t$.
		\For{$s=0$, $1$, $\dots$, $S-1$}\label{line:inner_loop_begin}
		\State Let $G_{s,t}(\cdot)=\lag(\cdot,\yB_t,\beta_t)+L_t\|\cdot-\xB_{s,t}\|^2$.
		\State $\xB_{s+1,t}=\bcd\big(G_{s,t}, h,\xB_{s,t}, 3L_t,\frac{\epsilon}{4}\big)$.
		\If{$2L_t\|\xB_{s+1,t}-\xB_{s,t}\|\le\frac{\epsilon}{2}$}
		\State $\xB_{t+1}=\xB_{s+1,t}$.
		\State \textbf{Break.}
		\EndIf
		\EndFor\label{line:inner_loop_end}
		\State $\yB_{t+1}=\yB_t+w_t\conB\big(\xB_{t}\big)$ where $w_t=w_0\min\{1,\frac{\log^2 2\|\conB(\xB_{1})\|}{(t+1)\log^2(t+2)\|\conB(\xB_{t+1})\|}\}$
		\EndFor
	\end{algorithmic}
\end{algorithm}

\paragraph{BCD.}
Each BCD iteration updates one randomly selected matrix using the partial gradient of $G(\cdot)$.
Other two matrices are fixed.
Note that all BCD iterations admit closed-form solutions, since the indicator functions and the $\ell_1$ norm regularizer are \emph{proximal-friendly}.

\begin{algorithm}[ht]
	\caption{Block coordinate descent method: $\bcd(G, h, \xB_0, L, \delta)$\label{alg:bcd}}
	\begin{algorithmic}[1]
		\State {\textbf{Input:}} $\xB_0\in\XM$, smoothness $L$, and stationary tolerance $\delta$.
		\For{$\tau=0,1,\dots$}
		\State Uniformly choose $i_\tau\in\{0,1,2\}$.
		\If{$i_\tau=0$}
		\State $\AB_{\tau+1}=\argmin_\AB\langle\nabla_\AB G(\xB_\tau),\AB\rangle+\frac{L}{2}\|\AB-\AB_\tau\|^2_F+h(\AB,\BB_\tau,\SB_{\tau})$, $\BB_{\tau+1}=\BB_\tau$, $\SB_{\tau+1}=\SB_{\tau}$.
		\ElsIf{$i_\tau=1$}
		\State $\BB_{\tau+1}=\argmin_\BB\langle\nabla_\BB G(\xB_\tau),\AB\rangle+\frac{L}{2}\|\BB-\BB_\tau\|^2_F+h(\AB_\tau,\BB,\SB_{\tau})$, $\AB_{\tau+1}=\AB_\tau$, $\SB_{\tau+1}=\SB_{\tau}$.
		\Else
		\State $\SB_{\tau+1}=\argmin_\SB\langle\nabla_\SB G(\xB_\tau),\SB\rangle+\frac{L}{2}\|\SB-\SB_\tau\|^2_F+h(\AB_\tau,\BB_{\tau},\SB)$, $\AB_{\tau+1}=\AB_\tau$, $\BB_{\tau+1}=\BB_{\tau}$.
		\EndIf
		\If{$\dist\big(-\nabla G(\xB_{\tau+1}),\partial h(\xB_{\tau+1})\big)\le\delta$}
		\State \textbf{Return} $\xB_{\tau+1}=\big(\vecto(\AB_{\tau+1});\vecto(\BB_{\tau+1});\vecto(\SB_{\tau+1})\big)$
		\EndIf
		\EndFor
	\end{algorithmic}
\end{algorithm}

\subsection{Complexity Analysis}\label{sec:complexity}
We first bound the number of BCD iterations that is required to reach a stationary point of (\ref{eq:problem_alm}).
The computational cost for each BCD iteration is then analyzed.
Finally, we obtain the overall complexity of Algorithm \ref{alg:ialm}.
For simplicity, we assume $m\le n$ without loss of generality in this subsection.
Detailed proofs are provided in the appendix.

\paragraph{Number of BCD iterations.}
Following the literature \citep{sahin2019inexact,li2021rate}, we analyze the complexity for reaching a first-order stationary point which is defined as follows.
\begin{definition}
	A pair $(\xB,\yB)$ is called an $\epsilon$-KKT point to (\ref{eq:problem_alm}) \citep{sahin2019inexact,li2021rate} if
	\begin{equation}
		\sqrt{\sum_{u=0}^{m+n}\con_u(\xB)^2}\le\epsilon,
	\end{equation}
	and
	\begin{equation}
		\dist\big(-\nabla_\xB \lag(\xB,\yB,\beta),\partial h(\xB)\big)\le\epsilon,
	\end{equation}
	hold,
	where the distance function between a vector $\aB$ and a set $\BM$ is defined as $\dist(\aB,\BM)=\min_{\bB\in\BM}\|\aB-\bB\|$.
\end{definition}
The main convergence result is summarized in the following theorem.
\begin{restatable}{proposition}{proptotaliter}\label{prop:total_iter}
	In order to produce an $\epsilon$-KKT solution of (\ref{eq:problem_alm}) 
	Algorithm \ref{alg:ialm} updates $\AB$, $\BB$, and $\SB$ for $\OM\Big(\frac{1}{\epsilon^3}(\log\frac{1}{\epsilon})^2\Big)$ times in expectation.
\end{restatable}
The number of iterations is the same as the Sinkhorn method in terms of $\epsilon$ up to logarithm factors.

\paragraph{Per-iteration complexity.}
The cost matrix $\CB$ is often low-rank, which can be exploited to accelerate the computation of BCD iterations \citep{scetbon2021low}.
Under suitable assumptions, the per-iteration complexity for each BCD iteration is $\OM(nr^2)$, which is stated in the following proposition.

\begin{restatable}{proposition}{propperiter}\label{prop:per_iter}
	When the following assumptions hold,
	\begin{enumerate}
		\item $\|\SB_\tau\|_0\le \rho'$ for all $\tau$ where $\rho'=\OM(nr)$,
		\item and $r\ge\rank(\CB)$,
	\end{enumerate}
	the per-iteration complexity for each BCD iteration is $\OM(nr^2)$ by exploiting the low-rank structure of $\CB$ in evaluating the partial gradients.
\end{restatable}
The first assumption is mild by choosing a moderately large $\lambda$.


%% file: sections/proof.tex
\subsection{Miscellaneous Helpful Lemmas}

\begin{lemma}[Lemma K of \citet{altschuler2019massively}]\label{lem:log_diff}
	For any $a,b>0$,
	\begin{equation*}
		|\log a-\log b|\le \frac{|a-b|}{\min\{a,b\}}.
	\end{equation*}
\end{lemma}

\begin{lemma}[Proposition 1.3 of \citet{bubeck2015convex}]\label{lem:first_order}
	Let $f$ be convex and $\XM$ a closed convex set on which $f$ is differentiable.
	Then
	\begin{equation*}
		\xB^*\in\argmin_{\xB\in\XM}f(\xB),
	\end{equation*}
	if and only if one has
	\begin{equation*}
		\langle \nabla f(\xB^*),\xB^*-\yB\rangle \le 0, \forall \yB\in\XM.
	\end{equation*}
\end{lemma}

\begin{lemma}[Sinkhorn projection]\label{lem:sinkhorn}
	Given $\pB\in\Delta^m$, $\qB\in\Delta^n$ and $\XB\in\RBB_+^{m\times n}$, the \emph{Sinkhorn projection} $\proj(\XB)$ of $\XB$ onto $\Pi(\pB,\qB)$ defined as
	\begin{equation}\label{eq:sinkhorn_projection}
		\proj(\XB)=\argmin_{\TB\in\Pi(\pB,\qB)}\KL(\TB\|\XB),
	\end{equation}
	is the unique matrix in $\Pi(\pB,\qB)$ of the form $\DB_1\KB\DB_2$ where $\DB_1$ and $\DB_2$ are diagonal matrices with strictly positive diagonal elements.
\end{lemma}

\noindent{\bf Proof:}

The strict convexity of KL-divergence and the compactness of $\Pi(\pB,\qB)$ implies that the minimizer exists and is unique.
Introducing two dual variables $\fB\in\RBB^m$, $\gB\in\RBB^n$ for each marginal constraint, the Lagrangian of Eq. (\ref{eq:sinkhorn_projection}) reads
\begin{equation*}
	\textgoth{L}(\TB,\fB,\gB)=\sum_{ij}T_{ij}\Big(\log\frac{T_{ij}}{X_{ij}}-1\Big)+\langle \fB,\TB\oneB-\pB\rangle + \langle \gB, \TB^\top\oneB-\qB\rangle.
\end{equation*}
First order conditions then yield
\begin{equation*}
	\frac{\partial \textgoth{L}}{\partial T_{ij}} = \log\frac{T_{ij}}{X_{ij}}+f_i+g_j=0,
\end{equation*}
which result in the expression
\begin{equation*}
	\proj(\XB) = \diag\big(\exp(-\fB)\big)\XB\diag\big(\exp(-\gB)\big).
\end{equation*}
\begin{flushright}
	$\blacksquare$
\end{flushright}

\subsection{Missing proofs in Sec. \ref{sec:approximated_distance}}

We first list some lemmas which are useful to prove Theorem \ref{thm:ls}.

\begin{lemma}\label{lem:llss}
	Under assumptions of Theorem \ref{thm:ls},
	\begin{equation*}
		\|\LB^*+\SB^*-\tilde{\LB^*}-\tilde{\SB^*}\|_\infty \le U[(r^*-r)_+\sqrt{mn}+(\rho^*-\rho)_+].
	\end{equation*}
\end{lemma}

\noindent{\bf Proof:}

When $r\ge r^*$ (resp. $\rho\ge\rho^*$), $\tilde{\LB^*}$ (resp. $\tilde{\SB^*}$) can accurately recover $\LB^*$ (resp. $\SB^*$).
$\tilde{\SB^*}$ is the best approximation for $\SB^*$ with at most $\rho$ \nonzero{} entries, which implies
\begin{equation}\label{eq:ss_Frob}
	\|\SB^*-\tilde{\SB^*}\|_F\le(\rho^*-\rho)_+U.
\end{equation}
By the definition of the \nonnegative{} rank, there exists additive decomposition
\begin{equation*}
	\LB^*=\sum_{i=0}^{r^*-1}\RB_i, \text{ s.t. }\rank(\RB_i)=1,\RB_i\ge 0.
\end{equation*}
When $r<r*$,
\begin{equation*}
	\|\tilde{\LB^*}-\LB^*\|_F \le \|\sum_{i=r}^{r^*-1}\RB_i\|_F \le \sum_{i=r}^{r^*-1}\|\RB_i\|_F \le (r^*-r)\|\LB^*\|_F \le \sqrt{mn}(r^*-r)U,
\end{equation*}
where the four inequalities are due to the definition of $\tilde{\LB^*}$, the definition of the matrix norm, the \nonnegativity{} of $\RB_i$'s, and the relation between the Frobenius norm and the infinity norm.
We then have
\begin{equation}\label{eq:ll_Frob}
	\|\tilde{\LB^*}-\LB^*\|_F \le \sqrt{mn}(r^*-r)_+U.
\end{equation}
Combining Eq. (\ref{eq:ss_Frob}) and (\ref{eq:ll_Frob}), we have
\begin{equation*}
	\|\LB^*+\SB^*-\tilde{\LB^*}-\tilde{\SB^*}\|_\infty \le \|\LB^*-\tilde{\LB^*}\|_\infty+\|\SB^*-\tilde{\SB^*}\|_\infty \le \|\LB^*-\tilde{\LB^*}\|_F+\|\SB^*-\tilde{\SB^*}\|_F \le U[(r^*-r)_+\sqrt{mn}+(\rho^*-\rho)_+].
\end{equation*}
\begin{flushright}
	$\blacksquare$
\end{flushright}

\noindent We now define an auxiliary function which is necessary to prove Theorem \ref{thm:ls}, i.e.,
\begin{equation*}
	\psi(x)=\begin{cases}
		\log x, &\text{ if }x>0,\\
		\log \delta, &\text{ otherwise,}
	\end{cases}
\end{equation*}
where $\delta$ is defined as Eq. (\ref{eq:delta}).
With slight abuse of notation, $\psi(\XB)$ is the \elementwise{} operation for matrix $\XB$.

\begin{lemma}\label{lem:psi}
	Let $\tilde{\TB}=\proj(\tilde{\LB^*}+\tilde{\SB^*})$.
	Under assumptions of Theorem \ref{thm:ls},
	\begin{equation}\label{eq:psi}
		\|\tilde{\TB}-\TB^*\|_1\le \|\psi(\tilde{\LB^*}+\tilde{\SB^*})-\psi(\LB^*+\SB^*)\|_\infty.
	\end{equation}
\end{lemma}
\noindent{\bf Proof:}

The case where $r\ge r^*$ and $\rho\ge \rho^*$ is obvious since $\LB^*$ and $\SB^*$ can be accurately recovered.

Now we consider the case where $r<r^*$ or $\rho<\rho^*$ holds, or both hold.
For notational simplicity, let $\tilde{\ZB}=\tilde{\LB^*}+\tilde{\SB^*}$.
By Lemma \ref{lem:first_order} and the form of Sinkhorn projection, $\sum_{ij}\log\frac{\tilde{T}_{ij}}{\tilde{Z}_{ij}}(T_{ij}^*-\tilde{T}_{ij})\ge 0$,
which leads to
\begin{equation*}
	\sum_{ij}\big(\psi(\tilde{T}_{ij})-\psi(\tilde{Z}_{ij})\big)(T_{ij}^*-\tilde{T}_{ij})\ge 0.
\end{equation*}
Since $\sum_{ij}\big(\psi(T^*_{ij})-\psi(T^*_{ij})\big)(T_{ij}^*-\tilde{T}_{ij})= 0$,
we have
\begin{equation*}
	\sum_{ij}\big(\psi(\tilde{T}_{ij})-\psi(\tilde{Z}_{ij})+\psi(T^*_{ij})-\psi(T^*_{ij})\big)(\tilde{T}_{ij}-T^*_{ij})\le 0,
\end{equation*}
which can be rearranged as
\begin{equation*}
	\sum_{ij}\big(\psi(\tilde{T}_{ij})-\psi(T^*_{ij})\big)(\tilde{T}_{ij}-T^*_{ij})\le \sum_{ij}\big(\psi(\tilde{Z}_{ij})-\psi(T^*_{ij})\big)(\tilde{T}_{ij}-T^*_{ij}) \le \|\psi(\tilde{\ZB})-\psi(\TB^*)\|_\infty\|\tilde{\TB}-\TB^*\|_1,
\end{equation*}
where we use H\"{o}lder's inequality for the second inequality.
To obtain Eq. (\ref{eq:psi}), it suffices to prove that
\begin{equation}\label{eq:first_order_holder}
	\|\tilde{\TB}-\TB^*\|_1^2\le \sum_{ij}\big(\psi(\tilde{T}_{ij})-\psi(T^*_{ij})\big)(\tilde{T}_{ij}-T^*_{ij}).
\end{equation}
$\psi(\tilde{T}_{ij})-\psi(T^*_{ij})$ has four possible forms
\begin{equation*}
	\psi(\tilde{T}_{ij})-\psi(T^*_{ij})=\begin{cases}
		\log \tilde{T}_{ij}-\log T^*_{ij},&\text{ if }\tilde{T}_{ij}>0 \text{ and } T_{ij}^*>0\\
		\log \tilde{T}_{ij}-\log \delta,&\text{ if }\tilde{T}_{ij}>0 \text{ and } T_{ij}^*=0\\
		\log \delta-\log T^*_{ij},&\text{ if }\tilde{T}_{ij}=0 \text{ and } T_{ij}^*>0\\
		\log \delta - \log \delta, &\text{ otherwise}
	\end{cases},
\end{equation*}
which all lead to $\big(\psi(\tilde{T}_{ij})-\psi(T^*_{ij})\big)(\tilde{T}_{ij}-T^*_{ij})\ge 0$.
Let $y_{ij}=\big(\psi(\tilde{T}_{ij})-\psi(T^*_{ij})\big)(\tilde{T}_{ij}-T^*_{ij})$.
Denote $\IM=\{(i,j)|y_{ij}>0\}$.
Based
on the Cauchy-Schwartz inequality, we can bound the left-hand side of Eq. (\ref{eq:first_order_holder}) as follows
\begin{equation*}
	\|\tilde{\TB}-\TB^*\|_1^2=\bigg(\sum_{(i,j)\in\IM}\sqrt{y_{ij}}\frac{|\tilde{T}_{ij}-T^*_{ij}|}{\sqrt{y_{ij}}}\bigg)^2 \le \sum_{(i,j)\in\IM}y_{ij}\sum_{(i,j)\in\IM} \frac{(\tilde{T}_{ij}-T^*_{ij})^2}{y_{ij}}.
\end{equation*}
We thus prove $\sum_{(i,j)\in\IM}\frac{(\tilde{T}_{ij}-T^*_{ij})^2}{y_{ij}}\le 1$, i.e.,
\begin{equation}\label{eq:sum_le_1}
	\sum_{(i,j)\in\IM}\frac{\tilde{T}_{ij}-T^*_{ij}}{\psi(\tilde{T}_{ij})-\psi(T^*_{ij})}\le 1.
\end{equation}
To do so, we show in the sequel that for all $(i,j)\in\IM$,
\begin{equation}\label{eq:magic}
	\frac{\tilde{T}_{ij}-T^*_{ij}}{\psi(\tilde{T}_{ij})-\psi(T^*_{ij})} \le \frac{\tilde{T}_{ij}+T^*_{ij}}{2},
\end{equation}
which can immediately imply Eq. (\ref{eq:sum_le_1}).
The left-hand side of the above is positive and thus we can assume without loss of generality $\tilde{T}_{ij}>T^*_{ij}$.
We consider separately the cases $T^*_{ij}>0$ and $T^*_{ij}=0$.

\noindent{\bf (i)} $T^*_{ij}>0$.
Fix $T^*_{ij}$ and consider the function
\begin{equation*}
	\phi(x)=2(x-T^*_{ij})-(x+T^*_{ij})\big(\psi(x)-\psi(T^*_{ij})\big).
\end{equation*}
We now prove $\phi(x)\le 0$ for $x\ge T^*_{ij}$.
Clearly $\phi(T^*_{ij})=0$ and
\begin{equation*}
	\phi(x)=2(x-T^*_{ij})-(x+T^*_{ij})\big(\log x-\log T^*_{ij}\big).
\end{equation*}
Its derivative is negative,
\begin{equation*}
	\phi'(x)=2-(\log x-\log T^*_{ij})-(x+T^*_{ij})\frac{1}{x}=1+\log\frac{T^*_{ij}}{x}-\frac{T^*_{ij}}{x} \le 0,
\end{equation*}
where in the last inequality we use $1+\log a\le a$.
We hence have the result.

\noindent{\bf (ii)} $T^*_{ij}>0$.
The left-hand side of Eq. (\ref{eq:magic}) can be bounded as follows
\begin{equation*}
	\frac{\tilde{T}_{ij}-T^*_{ij}}{\psi(\tilde{T}_{ij})-\psi(T^*_{ij})} = \frac{\tilde{T}_{ij}}{\log\tilde{T}_{ij}-\log\delta} \le \frac{\tilde{T}_{ij}}{2},
\end{equation*}
where in the last inequality we use the definition of $\delta$.

\noindent Combining the two cases above , $\frac{\tilde{T}_{ij}-T^*_{ij}}{\psi(\tilde{T}_{ij})-\psi(T^*_{ij})} \le \frac{\tilde{T}_{ij}+T^*_{ij}}{2}$ and $\|\tilde{\TB}-\TB^*\|_1^2\le \sum_{ij}\big(\psi(\tilde{T}_{ij})-\psi(T^*_{ij})\big)(\tilde{T}_{ij}-T^*_{ij})$, which finishes the proof.

\begin{flushright}
	$\blacksquare$
\end{flushright}

\noindent Now we prove Theorem \ref{thm:ls} which is restated here for convenience.
\primetheorem*

\noindent{\bf Proof:}

Eq. (\ref{eq:ls}) is obvious for $r\ge r^*$ and $\rho\ge\rho^*$.

Now we consider the case where $r<r^*$ or $\rho<\rho^*$ holds, or both hold.
Let $\tilde{\TB}=\proj(\tilde{\LB^*}+\tilde{\SB^*})$.
By the property of \nonnegative{} rank and the form of Sinkhorn projection, $\tilde{\TB}\in\tilde{\Omega}_{r,\rho}(\pB,\qB)$.
We thereby have
\begin{equation}\label{eq:whole}
	\langle \CB,\hat{\TB}\rangle-\langle\CB,\TB^*\rangle \le \langle\CB,\tilde{\TB}\rangle-\langle\CB,\TB^*\rangle \le \|\CB\|_\infty\|\tilde{\TB}-\TB^*\|_1 \le \|\CB\|_\infty\|\psi(\tilde{\LB^*}+\tilde{\SB^*})-\psi(\LB^*+\SB^*)\|_\infty,
\end{equation}
where we use H\"{o}lder's inequality  and Lemma \ref{lem:psi} for the third and the fourth inequalities respectively.
Now we bound $\|\psi(\tilde{\LB^*}+\tilde{\SB^*})-\psi(\LB^*+\SB^*)\|_\infty$ and consider the four following cases.

\noindent{\bf (i)} When $\tilde{L^*_{ij}}+\tilde{S^*_{ij}}>0$ and $L^*_{ij}+S^*_{ij}>0$,
\begin{equation*}
	\Big|\psi(\tilde{L^*_{ij}}+\tilde{S^*_{ij}})-\psi(L^*_{ij}+S^*_{ij})\Big| = \Big|\log(\tilde{L^*_{ij}}+\tilde{S^*_{ij}})-\log(L^*_{ij}+S^*_{ij})\Big| \le \frac{\Big|\tilde{L^*_{ij}}+\tilde{S^*_{ij}}-L^*_{ij}-S^*_{ij}\Big|}{\min\{\tilde{L^*_{ij}}+\tilde{S^*_{ij}},L^*_{ij}+S^*_{ij}\}} \le \frac{\Big|\tilde{L^*_{ij}}+\tilde{S^*_{ij}}-L^*_{ij}-S^*_{ij}\Big|}{\delta},
\end{equation*}
where we apply Lemma \ref{lem:log_diff} in the first inequality, and use the definition of $\delta$ in the second one.

\noindent{\bf (ii)} When $\tilde{L^*_{ij}}+\tilde{S^*_{ij}}>0$ and $L^*_{ij}+S^*_{ij}=0$,
\begin{equation*}
	\Big|\psi(\tilde{L^*_{ij}}+\tilde{S^*_{ij}})-\psi(L^*_{ij}+S^*_{ij})\Big| = \Big|\log(\tilde{L^*_{ij}}+\tilde{S^*_{ij}})-\log(\delta)\Big| \le \frac{\Big|\tilde{L^*_{ij}}+\tilde{S^*_{ij}}-\delta\Big|}{\min\{\tilde{L^*_{ij}}+\tilde{S^*_{ij}},\delta\}} \le \frac{\Big|\tilde{L^*_{ij}}+\tilde{S^*_{ij}}-L^*_{ij}-S^*_{ij}\Big|}{\delta},
\end{equation*}
where we use Lemma \ref{lem:log_diff} and the fact that $L^*_{ij}+S^*_{ij}=0$ in the first and the second inequalities respectively.

\noindent{\bf (iii)} When $\tilde{L^*_{ij}}+\tilde{S^*_{ij}}=0$ and $L^*_{ij}+S^*_{ij}>0$, we similarly have
\begin{equation*}
	\Big|\psi(\tilde{L^*_{ij}}+\tilde{S^*_{ij}})-\psi(L^*_{ij}+S^*_{ij})\Big| \le \frac{\Big|\tilde{L^*_{ij}}+\tilde{S^*_{ij}}-L^*_{ij}-S^*_{ij}\Big|}{\delta}.
\end{equation*}

\noindent{\bf (iv)} When $\tilde{L^*_{ij}}+\tilde{S^*_{ij}}=0$ and $L^*_{ij}+S^*_{ij}=0$, we have
\begin{equation*}
	\Big|\psi(\tilde{L^*_{ij}}+\tilde{S^*_{ij}})-\psi(L^*_{ij}+S^*_{ij})\Big| = \Big|\log(\delta)-\log(\delta)\Big| = \frac{\Big|\tilde{L^*_{ij}}+\tilde{S^*_{ij}}-L^*_{ij}-S^*_{ij}\Big|}{\delta}.
\end{equation*}

\noindent Combining the four cases, we have
\begin{equation}\label{eq:llss}
	\|\psi(\tilde{\LB^*}+\tilde{\SB^*})-\psi(\LB^*+\SB^*)\|_\infty \le \frac{1}{\delta} \|\LB^*+\SB^*-\tilde{\LB^*}-\tilde{\SB^*}\|_\infty \le \frac{U}{\delta}[(r^*-r)_+\sqrt{mn}+(\rho^*-\rho)_+],
\end{equation}
where we apply Lemma \ref{lem:llss} in the second inequality.
Substituting Eq. (\ref{eq:llss}) into (\ref{eq:whole}), we have the result.
\begin{flushright}
	$\blacksquare$
\end{flushright}

\subsection{Missing Proofs in Sec. \ref{sec:optimization}}
Proposition \ref{prop:smooth_sc} is restated here.

\propsmoothsc*

\noindent{\bf Proof:}

Recalling the definition $\xB=\big(\vecto(\AB);\vecto(\BB);\vecto(\SB)\big)$, it suffices to prove that $\lag(\xB,\yB,\beta)$ is $L(\yB,\beta)$-smooth.
We proceed by bounding the eigenvalues of the Hessian $\nabla^2\lag(\xB,\yB,\beta)$ given by
\begin{equation*}
	\nabla^2\lag(\xB,\yB,\beta)=\nabla^2\langle \CB,\AB\BB^\top+\SB\rangle+\nabla^2\sum_u\beta\big(\con_u(\xB)\big)^2+\nabla^2\sum_uy_u\con_u(\xB).
\end{equation*}
The Hessian of $\langle \CB,\AB\BB^\top+\SB\rangle$ is given by $\HB=[H_{lz}]$, where
\begin{equation*}
	H_{lz}=\begin{cases}
		C_{l\quo r,z\quo r-m}, & \text{ if }0\le l<mr, mr\le z<mr+nr, \text{ and }l\rem r= z\rem r\\
		C_{z\quo r,l\quo r-m}, & \text{ if }0\le z<mr, mr\le l<mr+nr, \text{ and }l\rem r= z\rem r\\
		0, & \text{ otherwise}
	\end{cases},
\end{equation*}
where $\quo$ and $\rem$ is the operation of obtaining the quotient and the remainder of the Euclidean division respectively.

The summands of the second term and the third term are
\begin{equation*}
	\nabla^2 \beta\big(\con_u(\xB)\big)^2 = \beta \con_u(\xB)\nabla^2\con_u(\xB) + \beta\nabla\con_u(\xB)\nabla^\top\con_u(\xB),
\end{equation*}
and
\begin{equation*}
	\nabla^2 y_u\con_u(\xB)=y_u\nabla^2\con_u(\xB),
\end{equation*}
respectively both of which involve Hessians $\HB^{\con_u}=[H^{\con_u}_{lz}]$, where for $0\le u<m$,
\begin{equation*}
	H^{\con_u}_{lz}=\begin{cases}
		1, & \text{ if }0\le l<mr, mr\le z<mr+nr, l\rem r= z\rem r, \text{ and }i=l\rem r\\
		1, & \text{ if }0\le z<mr, mr\le l<mr+nr, l\rem r= z\rem r, \text{ and }i=z\rem r\\
		0, & \text{ otherwise}
	\end{cases},
\end{equation*}
and for $m\le u<m+n-1$,
\begin{equation*}
	H^{\con_u}_{lz}=\begin{cases}
		1, & \text{ if }0\le l<mr, mr\le z<mr+nr, l\rem r= z\rem r, \text{ and }j=z\quo r - m\\
		1, & \text{ if }0\le z<mr, mr\le l<mr+nr, l\rem r= z\rem r, \text{ and }j=l\quo r-m\\
		0, & \text{ otherwise}
	\end{cases},
\end{equation*}
Therefore,
\begin{equation*}
	\begin{aligned}
		\|\nabla^2\lag(\xB,\yB,\beta)\|_{\op} &\le \|\nabla^2\lag(\xB,\yB,\beta)\|_{F}\\
		&\le \|\HB\|_F+\beta\sum_{u=0}^{m+n-1}|\con_u(\xB)|\|\HB^{\con_u}\|_F+\beta\sum_{u=0}^{m+n-1}\|\nabla\con_u(\xB)\nabla^\top\con_u(\xB)\|_F^2+\sum_{u=0}^{m+n-1}|y_u|\|\HB^{\con_u}\|_F\\
		&\le \sqrt{2r}\|\CB\|_F+\beta\sum_{u=0}^{m-1}B_u\sqrt{2nr}+\beta\sum_{u=m}^{m+n-1}B_u\sqrt{2mr}+\beta\sum_{u=0}^{m+n-1}B_u^2+\sqrt{2r}(m\sqrt{n}+n\sqrt{m})\|\yB\|,
	\end{aligned}
\end{equation*}
which indicates that $\lag(\xB,\yB,\beta)$ is $L(\yB,\beta)$-smooth.
\subsection{Missing Proofs in Sec. \ref{sec:complexity}}

\begin{lemma}[Complexity of BCD]\label{lem:bcd}
	Given $\epsilon>0$, within $\OM\big(\log(\frac{1}{\epsilon})\big)$ iterations in expectation, Algorithm \ref{alg:bcd} outputs a solution $\xB$ that satisfies $\dist\big(-\nabla G(\xB),\partial h(\xB)\big)\le\epsilon$.
\end{lemma}

\noindent{\bf Proof:}
For ease of notation, we denote $F(\xB)=G(\xB)+h(\xB)$.
By Proposition \ref{prop:smooth_sc} and Theorem 7 of \citet{richtarik2014iteration}, we have
\begin{equation}\label{eq:linear_convergence_bcd}
	\EBB_\tau F(\xB_T)-F(\xB^*) \le \big(\frac{8}{9}\big)^T\big(F(\xB_0)-F(\xB^*) \big),
\end{equation}
where $\xB^*=\argmin_\xB F(\xB)$.
By the $3L(\yB,\beta)$-smoothness of $F(\xB)$, we have
\begin{equation}\label{eq:smoothness_lower}
	F(\xB_T)-F(\xB^*) \ge \frac{1}{2\cdot 3L(\yB,\beta)}\|\gB(\xB)\|^2.
\end{equation}
Combining (\ref{eq:linear_convergence_bcd}) and (\ref{eq:smoothness_lower}), we have $T=\OM\Big(\log\frac{L(\yB,\beta)(F(\xB_0)-F(\xB^*))}{\epsilon^2}\Big)$.

\begin{flushright}
	$\blacksquare$
\end{flushright}


\proptotaliter*

\noindent{\bf Proof:}

Invoking Theorem 2 of \citet{li2021rate}, Algorithm \ref{alg:ialm} terminates with $T=\OM(\log\frac{1}{\epsilon})$ and $S=\OM(\frac{1}{\epsilon^3})$.
By Lemma \ref{lem:bcd}, Algorithm \ref{alg:bcd} updates $\AB$, $\BB$, and $\SB$ for $\OM(\log\frac{1}{\epsilon})$ times in expectation.
We hence have the results.

\begin{flushright}
	$\blacksquare$
\end{flushright}

\propperiter*

\noindent{\bf Proof:}

Denote the operation of clamping each entry of matrix $\XB$ to a box $[l, u]$ by $\clamp(\XB;l,u)$ and the shrinkage operator by $\shrin(\XB,a)$.
Then
\begin{subequations}
	\begin{alignat}{3}
		\AB_{\tau+1}&=\clamp\big(\AB_\tau-\frac{1}{L}\nabla_\AB G(\xB_\tau),0,1\big),\\
		\BB_{\tau+1}&=\clamp\big(\BB_\tau-\frac{1}{L}\nabla_\BB G(\xB_\tau),0,1\big),\\
		\SB_{\tau+1}&=\clamp\Big(\shrin\big(\SB-\frac{1}{L}\nabla_\SB G(\xB_\tau),\frac{\lambda}{L}\big),0,1\Big),\label{eq:update_S}
	\end{alignat}
\end{subequations}
where
\begin{equation*}
	\begin{aligned}
		\nabla_\AB G(\xB_\tau)&=\CB\BB_{\tau}+\yp{\oneB^n}^\top\BB_{\tau}+\oneB^m{\yq}^\top\BB_{\tau}+\beta\Big(\AB_\tau\BB_\tau^\top\oneB^n{\oneB^n}^\top\BB_{\tau}+\SB_\tau\oneB^n{\oneB^n}^\top\BB_\tau-\pB{\oneB^n}^\top\BB_{\tau}\Big)\\
		&+\beta\Big(\onem\onem^\top\AB_\tau\BB_\tau^\top\BB_{\tau}+\onem\onem^\top\SB_{\tau}\BB_{\tau}-\onem\yq^\top\BB_{\tau}\Big)+2L(\AB_\tau-\AB_0)\\
		\nabla_\BB G(\xB_\tau)&=\CB^\top\AB_\tau+\onen\yp^\top\AB_\tau+\yq\onem^\top\AB_\tau+\beta\Big(\onen\onen^\top\BB_\tau\AB_\tau^\top\AB_\tau+\onen\onen^\top\SB_{\tau}^\top\AB_\tau-\onen\pB^\top\AB_\tau\Big)\\
		&+\beta\Big(\BB_\tau\AB_\tau^\top\onem\onem^\top\AB_\tau+\SB_{\tau}^\top\onem\onem^\top\AB_\tau-\qB\onem^\top\AB_\tau\Big)+2L(\BB_{\tau}-\BB_0)\\
		\nabla_\SB G(\xB_\tau)&=-\DB+\beta\Big(\AB_\tau\BB_\tau^\top\onen\onen^\top+\SB_\tau\onen\onen^\top\Big)+\beta\Big(\onem\onem^\top\AB_\tau\BB_{\tau}^\top+\onem\onem^\top\SB_{\tau}\Big)+\rho\SB_{\tau},
	\end{aligned}
\end{equation*}
and
\begin{equation*}
	\DB=-\Big(\CB+\yp\onen^\top+\onem\yq^\top-\beta\pB\onen^\top-\beta\onem\qB^\top-\rho\SB_0\Big).
\end{equation*}
Exploiting the low-rankness of $\CB$, the complexity for updating $\AB$ and $\BB$ is obviously $\OM(nr^2)$.
Substituting $\nabla_\SB G(\xB_\tau)$ into Eq. (\ref{eq:update_S}), we have
\begin{equation*}
	\SB_{\tau+1}=\clamp\bigg(\shrin\Big(\frac{1}{3}\SB_{\tau}+\frac{1}{L}\DB-\MB,\frac{\lambda}{L}\Big),0,1\bigg),
\end{equation*}
where $\MB=\frac{\beta}{L}\big(\AB_\tau\BB_\tau^\top\onen\onen^\top+\SB_\tau\onen\onen^\top\big)+\frac{\beta}{L}\big(\onem\onem^\top\AB_\tau\BB_{\tau}^\top+\onem\onem^\top\SB_{\tau}\big)$.
If $\lambda$ is larger than or equal to the $nr$\textsuperscript{th} largest entry in $\DB$, only $\OM(\|\SB_\tau\|_0+nr)$ entries of $\frac{1}{3}\SB_{\tau}+\frac{1}{L}\DB-\MB$ will be larger than $\frac{\lambda}{L}$.
Then only $\OM(\|\SB_\tau\|_0+nr)$ of $\MB$ need to be calculated, with complexity $\OM(\|\SB_\tau\|_0+nr)$.
\begin{flushright}
	$\blacksquare$
\end{flushright}

%% file: ls.bbl
\begin{thebibliography}{23}
\providecommand{\natexlab}[1]{#1}
\providecommand{\url}[1]{\texttt{#1}}
\expandafter\ifx\csname urlstyle\endcsname\relax
  \providecommand{\doi}[1]{doi: #1}\else
  \providecommand{\doi}{doi: \begingroup \urlstyle{rm}\Url}\fi

\bibitem[Altschuler et~al.(2017)Altschuler, Weed, and
  Rigollet]{altschuler2017near}
Jason Altschuler, Jonathan Weed, and Philippe Rigollet.
\newblock Near-linear time approximation algorithms for optimal transport via
  sinkhorn iteration.
\newblock \emph{Advances in Neural Information Processing Systems},
  2017:\penalty0 1965--1975, 2017.

\bibitem[Altschuler et~al.(2019)Altschuler, Bach, Rudi, and
  Niles-Weed]{altschuler2019massively}
Jason Altschuler, Francis Bach, Alessandro Rudi, and Jonathan Niles-Weed.
\newblock Massively scalable sinkhorn distances via the nystr{\"o}m method.
\newblock In \emph{Proceedings of the 33rd International Conference on Neural
  Information Processing Systems}, pages 4427--4437, 2019.

\bibitem[Altschuler and Boix-Adsera(2020)]{altschuler2020polynomial}
Jason~M Altschuler and Enric Boix-Adsera.
\newblock Polynomial-time algorithms for multimarginal optimal transport
  problems with decomposable structure.
\newblock \emph{arXiv preprint arXiv:2008.03006}, 2020.

\bibitem[Arjovsky et~al.(2017)Arjovsky, Chintala, and
  Bottou]{arjovsky2017wasserstein}
Martin Arjovsky, Soumith Chintala, and L{\'e}on Bottou.
\newblock Wasserstein generative adversarial networks.
\newblock In \emph{International conference on machine learning}, pages
  214--223. PMLR, 2017.

\bibitem[Brualdi(2006)]{brualdi2006combinatorial}
Richard~A Brualdi.
\newblock \emph{Combinatorial matrix classes}, volume~13.
\newblock Cambridge University Press, 2006.

\bibitem[Bubeck(2015)]{bubeck2015convex}
S{\'e}bastien Bubeck.
\newblock Convex optimization: Algorithms and complexity.
\newblock \emph{Foundations and Trends{\textregistered} in Machine Learning},
  8\penalty0 (3-4):\penalty0 231--357, 2015.

\bibitem[Cand{\`e}s et~al.(2011)Cand{\`e}s, Li, Ma, and
  Wright]{candes2011robust}
Emmanuel~J Cand{\`e}s, Xiaodong Li, Yi~Ma, and John Wright.
\newblock Robust principal component analysis?
\newblock \emph{Journal of the ACM (JACM)}, 58\penalty0 (3):\penalty0 1--37,
  2011.

\bibitem[Courty et~al.(2016)Courty, Flamary, Tuia, and
  Rakotomamonjy]{courty2016optimal}
Nicolas Courty, R{\'e}mi Flamary, Devis Tuia, and Alain Rakotomamonjy.
\newblock Optimal transport for domain adaptation.
\newblock \emph{IEEE transactions on pattern analysis and machine
  intelligence}, 39\penalty0 (9):\penalty0 1853--1865, 2016.

\bibitem[Cuturi(2013)]{cuturi2013sinkhorn}
Marco Cuturi.
\newblock Sinkhorn distances: Lightspeed computation of optimal transport.
\newblock \emph{Advances in neural information processing systems},
  26:\penalty0 2292--2300, 2013.

\bibitem[Forrow et~al.(2019)Forrow, H{\"u}tter, Nitzan, Rigollet, Schiebinger,
  and Weed]{forrow2019statistical}
Aden Forrow, Jan-Christian H{\"u}tter, Mor Nitzan, Philippe Rigollet, Geoffrey
  Schiebinger, and Jonathan Weed.
\newblock Statistical optimal transport via factored couplings.
\newblock In \emph{The 22nd International Conference on Artificial Intelligence
  and Statistics}, pages 2454--2465. PMLR, 2019.

\bibitem[Hestenes(1969)]{hestenes1969multiplier}
Magnus~R Hestenes.
\newblock Multiplier and gradient methods.
\newblock \emph{Journal of optimization theory and applications}, 4\penalty0
  (5):\penalty0 303--320, 1969.

\bibitem[Li et~al.(2021)Li, Chen, Liu, Lu, and Xu]{li2021rate}
Zichong Li, Pin-Yu Chen, Sijia Liu, Songtao Lu, and Yangyang Xu.
\newblock Rate-improved inexact augmented lagrangian method for constrained
  nonconvex optimization.
\newblock In \emph{International Conference on Artificial Intelligence and
  Statistics}, pages 2170--2178. PMLR, 2021.

\bibitem[Lin et~al.(2021)Lin, Azabou, and Dyer]{lin2021making}
Chi-Heng Lin, Mehdi Azabou, and Eva~L Dyer.
\newblock Making transport more robust and interpretable by moving data through
  a small number of anchor points.
\newblock \emph{Proceedings of machine learning research}, 139:\penalty0 6631,
  2021.

\bibitem[Lin et~al.(2011)Lin, Liu, and Su]{lin2011linearized}
Zhouchen Lin, Risheng Liu, and Zhixun Su.
\newblock Linearized alternating direction method with adaptive penalty for
  low-rank representation.
\newblock \emph{Advances in Neural Information Processing Systems},
  24:\penalty0 612--620, 2011.

\bibitem[Powell(1969)]{powell1969method}
Michael~JD Powell.
\newblock A method for nonlinear constraints in minimization problems.
\newblock \emph{Optimization}, pages 283--298, 1969.

\bibitem[Richt{\'a}rik and Tak{\'a}{\v{c}}(2014)]{richtarik2014iteration}
Peter Richt{\'a}rik and Martin Tak{\'a}{\v{c}}.
\newblock Iteration complexity of randomized block-coordinate descent methods
  for minimizing a composite function.
\newblock \emph{Mathematical Programming}, 144\penalty0 (1):\penalty0 1--38,
  2014.

\bibitem[Sahin et~al.(2019)Sahin, Alacaoglu, Latorre, Cevher,
  et~al.]{sahin2019inexact}
Mehmet~Fatih Sahin, Ahmet Alacaoglu, Fabian Latorre, Volkan Cevher, et~al.
\newblock An inexact augmented lagrangian framework for nonconvex optimization
  with nonlinear constraints.
\newblock \emph{Advances in Neural Information Processing Systems},
  32:\penalty0 13965--13977, 2019.

\bibitem[Scetbon and Cuturi(2020)]{scetbon2020linear}
Meyer Scetbon and Marco Cuturi.
\newblock Linear time sinkhorn divergences using positive features.
\newblock \emph{Advances in Neural Information Processing Systems}, 33, 2020.

\bibitem[Scetbon et~al.(2021)Scetbon, Cuturi, and Peyr{\'e}]{scetbon2021low}
Meyer Scetbon, Marco Cuturi, and Gabriel Peyr{\'e}.
\newblock Low-rank sinkhorn factorization.
\newblock In Marina Meila and Tong Zhang, editors, \emph{Proceedings of the
  38th International Conference on Machine Learning}, volume 139 of
  \emph{Proceedings of Machine Learning Research}, pages 9344--9354. PMLR,
  18--24 Jul 2021.
\newblock URL \url{https://proceedings.mlr.press/v139/scetbon21a.html}.

\bibitem[Tarjan(1997)]{tarjan1997dynamic}
Robert~E Tarjan.
\newblock Dynamic trees as search trees via euler tours, applied to the network
  simplex algorithm.
\newblock \emph{Mathematical Programming}, 78\penalty0 (2):\penalty0 169--177,
  1997.

\bibitem[Villani(2009)]{villani2009optimal}
C{\'e}dric Villani.
\newblock \emph{Optimal transport: old and new}, volume 338.
\newblock Springer, 2009.

\bibitem[Wright et~al.(2009)Wright, Ganesh, Rao, Peng, and
  Ma]{wright2009robust}
John Wright, Arvind Ganesh, Shankar~R Rao, Yigang Peng, and Yi~Ma.
\newblock Robust principal component analysis: Exact recovery of corrupted
  low-rank matrices via convex optimization.
\newblock In \emph{NIPS}, volume~58, pages 289--298, 2009.

\bibitem[Xu et~al.(2019)Xu, Luo, Zha, and Duke]{xu2019gromov}
Hongteng Xu, Dixin Luo, Hongyuan Zha, and Lawrence~Carin Duke.
\newblock Gromov-wasserstein learning for graph matching and node embedding.
\newblock In \emph{International conference on machine learning}, pages
  6932--6941. PMLR, 2019.

\end{thebibliography}
